\documentclass[conference]{IEEEtran}
\IEEEoverridecommandlockouts
\usepackage{cite}
\usepackage{amsmath,amssymb,amsfonts}
\usepackage{algorithmic}
\usepackage{graphicx}
\usepackage{textcomp}
\usepackage{xcolor}
\usepackage{tabu}
\DeclareMathOperator*{\Attention}{Attention}
\DeclareMathOperator*{\onehot}{one\_hot}

\def\BibTeX{{\rm B\kern-.05em{\sc i\kern-.025em b}\kern-.08emB
    T\kern-.1667em\lower.7ex\hbox{E}\kern-.125emX}}
\begin{document}
\title{Generating Chinese Poetry from Images via Concrete and Abstract Information
\thanks{This work is supported by the National Key R\&D Program of China under contract No. 2017YFB1002201, the National Natural Science Fund for Distinguished Young Scholar (Grant No. 61625204), and partially supported by the Key Program of National Science Foundation of China (Grant No. 61836006).}
}

\author{\IEEEauthorblockN{Yusen Liu}
\IEEEauthorblockA{\textit{Sichuan University} \\
Chengdu, China \\
liuyusen96@gmail.com}
\and
\IEEEauthorblockN{Dayiheng Liu}
\IEEEauthorblockA{\textit{Sichuan University} \\
Chengdu, China \\
losinuris@gmail.com}
\and
\IEEEauthorblockN{Jiancheng Lv}
\IEEEauthorblockA{\textit{Sichuan University} \\
Chengdu, China \\
lvjiancheng@scu.edu.cn}
\and
\IEEEauthorblockN{Yongsheng Sang}\thanks{Correspondence to Yongsheng Sang.}
\IEEEauthorblockA{\textit{Sichuan University} \\
Chengdu, China \\
sangys@scu.edu.cn}
}

\maketitle
\begin{abstract}
In recent years, the automatic generation of classical Chinese poetry has made great progress. Besides focusing on improving the quality of the generated poetry, there is a new topic about generating poetry from an image. However, the existing methods for this topic still have the problem of topic drift and semantic inconsistency, and the image-poem pairs dataset is hard to be built when training these models. In this paper, we extract and integrate the Concrete and Abstract information from images to address those issues. We proposed an infilling-based Chinese poetry generation model which can infill the Concrete keywords into each line of poems in an explicit way, and an abstract information embedding to integrate the Abstract information into generated poems. In addition, we use non-parallel data during training and construct separate image datasets and poem datasets to train the different components in our framework. Both automatic and human evaluation results show that our approach can generate poems which have better consistency with images without losing the quality.
\end{abstract}

\section{Introduction}
\begin{figure}
\centering
\includegraphics[width=1\linewidth]{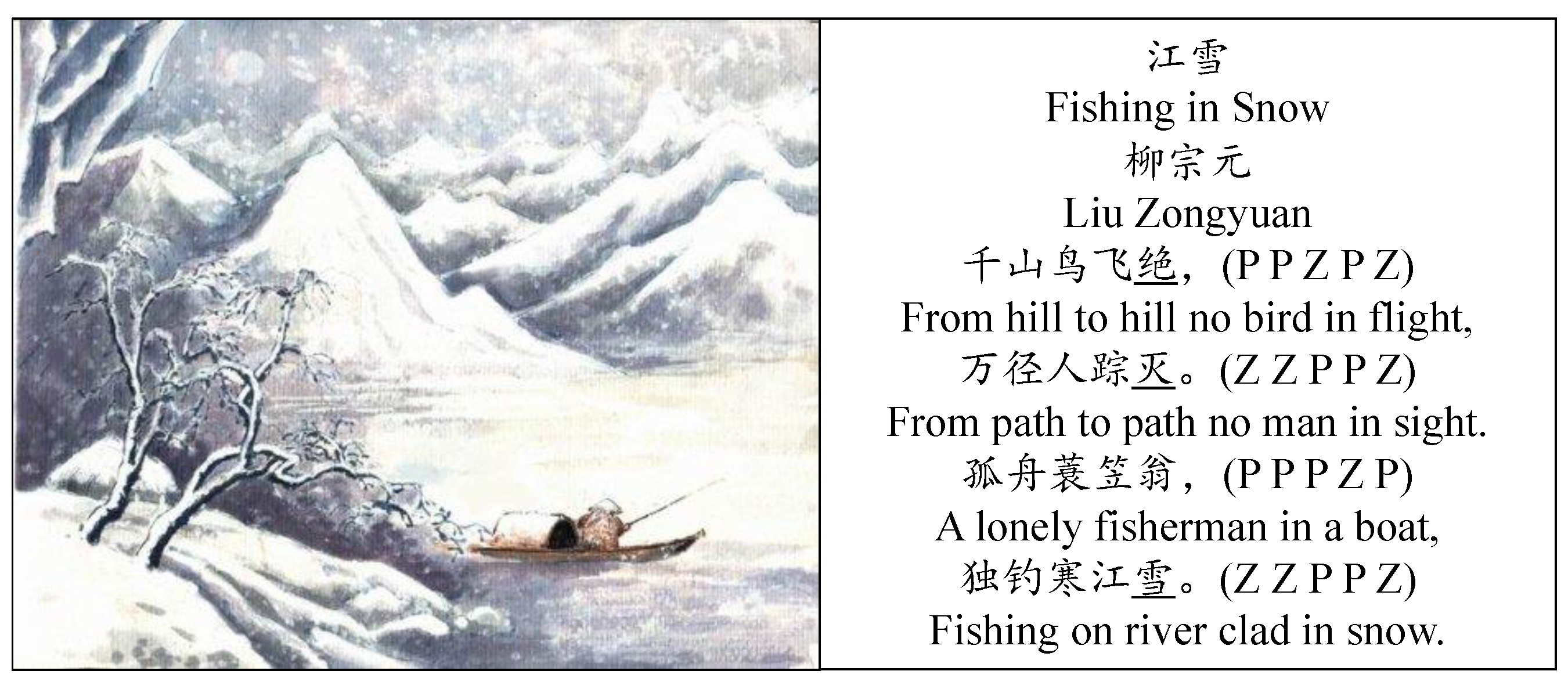}
\caption{An example of a 5-character \emph{quatrain}. The poet described the beautiful scenery shown in the left image of the figure. The rhyming characters are shown in underline. The tone of each character is shown at the end of each line, where ``P" and ``Z" are short-hands for Ping (the level tone) and Ze (the downward tone) respectively}
\label{fig1}
\end{figure}

Classical Chinese poetry is one of the great heritages in Chinese culture, as well as an important part of traditional Chinese literature. Among the various genres, \emph{quatrain} is the most popular one which consists of four lines and limits five or seven words each line. Rhyme and tone are essential for \emph{quatrain}. Rhyme means to put the same rhyme in the same place as each line, usually, the last characters in the first, second and fourth lines must have the same rhyme. In addition, each character has a particular tone, Ping (the level tone) or Ze (the downward tone). The Ping and Ze alternate in each line according to a certain rule, and they should correspond between each line\cite{wang2002summary}. An example of classical Chinese poetry is shown in Fig. \ref{fig1}. Generating Chinese classical poetry automatically is a meaningful and challenging research direction. There are several different attempts have been made on this task, such as genetic algorithms\cite{manurung2004evolutionary,zhou2010genetic,manurung2012using}, rule-based methods\cite{tosa2008hitch,wu2009new,netzer2009gaiku,oliveira2009automatic} and statistical machine translation methods\cite{jiang2008generating,he2012generating}. Furthermore, neural network have been applied in poetry generation \cite{zhang2014chinese,wang2016achinese,yi2017generating,wang2016bchinese,yi2018chinese}, these models can generate poetry according to several keywords or topics which given by users. 

However, in ancient China, the inspiration of ancient poets was not just several keywords or the figment of their imagination, many of them were inspired by beautiful scenery. A poet can describe a nature scene directly and then express his emotion, feelings or political views. For example, the poem shown in Fig. \ref{fig1} depicts a fisherman fishing alone in the snow. Seeing this, the poet takes advantage of the scene to express his disappointment in politics. Generally speaking, visual information is more natural and meaningful for writing poems. Therefore, Generating classical Chinese poetry from images is a reasonable and challenging task. There are few methods for generating poetry from images, such as generating Chinese poems based on the keywords extracted from images\cite{liu2018multi,cheng2018image} and generating poems from images in an end-to-end fashion\cite{xu2018images,liu2018beyond}. For these methods, there still have some imperfections need to be improved. The keyword-based methods cannot guarantee that the keywords will appear in the poem, which will lead to the problem of topic drift and semantic inconsistency. For the end-to-end methods, some important image information cannot be extracted accurately, such as seasonal information, yet such information is very essential for the generation model. For example, If we give the model a picture of spring, but the generated poem contains the phrases about winter, this will have a very bad influence on the consistency of poems and images. Moreover, these methods need the paired dataset of images and poems, but the dataset is hard to be built: the automatic matching method is very rough, and the manual method needs a great number of labor-power.

To tackle these problems, we first define the images information into two types: \textbf{Concrete} information and \textbf{Abstract} information, the concrete information is the concrete physical objects in images, such as mountains and rivers, the abstract information is the implied background information or people's common sense, such as season information. Then we propose a novel framework to extract these two kinds of information and generate poetry according to them. 

For \textbf{Concrete} information, we extract the keywords from an image as the concrete information and retrieve the extracted keywords from \emph{ShiXueHanYing}, which is an ancient Chinese phrase external knowledge base. Then we infill these keywords to each line of the generated poem using our infilling-based generation model, which bases on the Encoder-Decoder framework. Our method can ensure that the related keywords will appear in the poem. This solves the problem of topic drift and semantic inconsistency.

For \textbf{Abstract} information, we capture the abstract information from an image like season information using several image classification models. Then we propose the abstract information embedding to integrate this information with our model. This addresses the issue of some important image information cannot be extracted accurately, and improve the consistency of poems and images.

Especially, different from previous approaches, our method uses non-parallel data to train, which means that the image-poem paired data are not required. The experiments show that our method can better reflect the image information in poems and keep the high quality of the generated poems. Moreover, we collect separate image datasets and poem datasets, which are used to train the image classification models and then establish linkages between images features and poems contents without the paired dataset. The main contributions in this paper are concluded as follows: 

\begin{itemize}
\item To the best of our knowledge, this is the first method that generating poetry from images with non-parallel data.
\item We propose an infilling-based Chinese poetry generation model, which can ensure that the concrete keywords related to images will appear in the poems.
\item we propose the abstract information embedding to integrate the abstract information of images explicitly, which addresses the issue of some important image information cannot be extracted accurately.
\item we collect separate image datasets and poem datasets, which are used to establish linkages between images features and poems contents without the paired dataset. We will release these datasets soon.
\end{itemize}

\begin{figure*}
\centering
\includegraphics[width=0.8\textwidth]{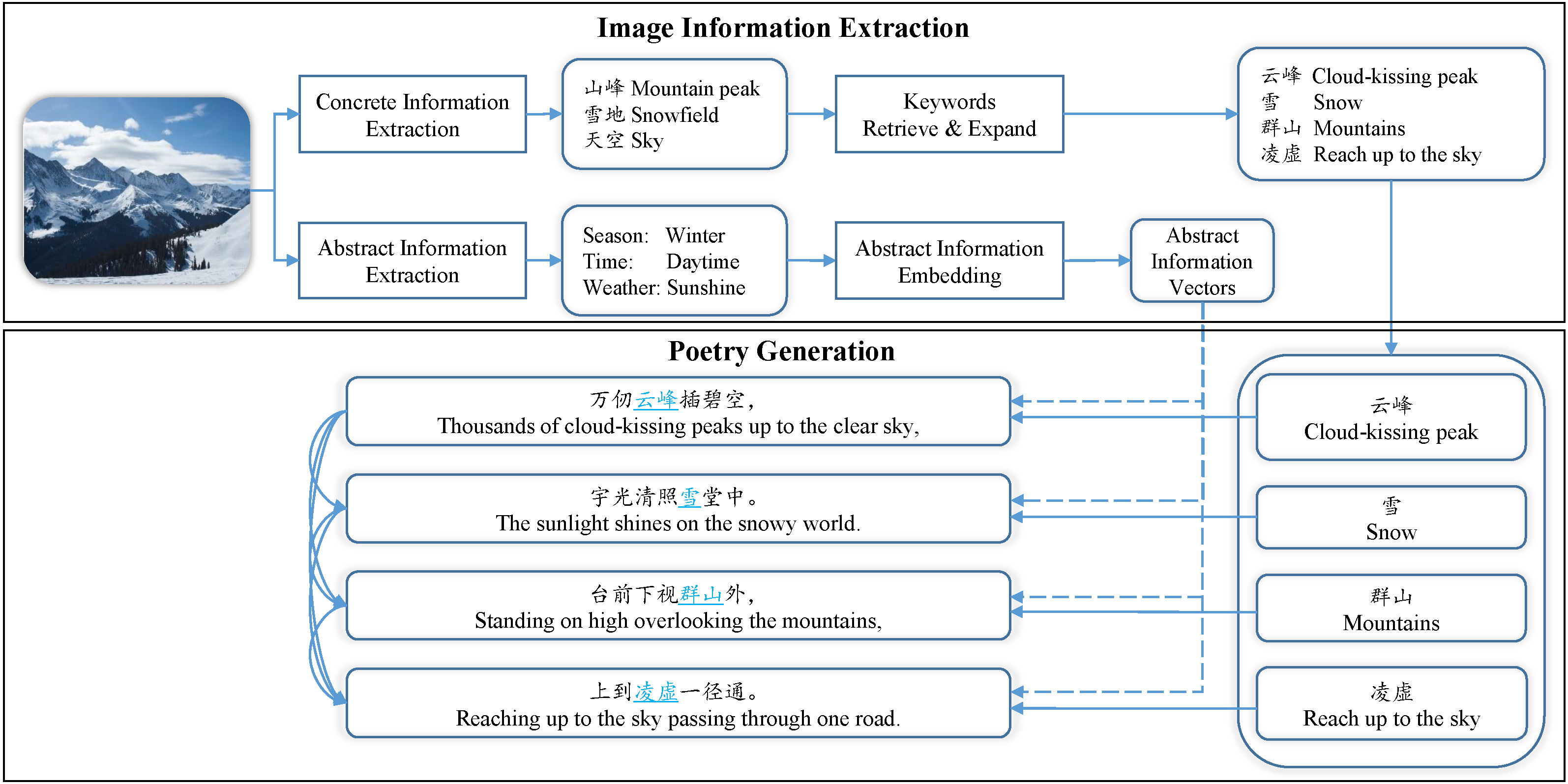}
\caption{An overview of our Chinese poetry generation framework.}
\label{fig2}
\end{figure*}

\section{Relate Work}
Poetry generation is a meaningful and challenging task. For the past decade, There are several approaches to address this issue. Some approaches are based on rules and templates. For instance, based on phrase searching\cite{tosa2008hitch,wu2009new,netzer2009gaiku}, semantic and grammar templates\cite{oliveira2009automatic}, genetic algorithms\cite{manurung2004evolutionary,manurung2012using,zhou2010genetic}. In these works, rules and templates are regarded as constraints or guidelines in the process of poetry generation. Furthermore, there is another approach which treats the poetry generation as a statistical machine translation problem\cite{jiang2008generating,he2012generating}.

With the development of deep learning, many new methods have been applied in poetry generation. Firstly, \cite{zhang2014chinese} proposes a model based on recurrent neural networks to generate classical Chinese poetry. The model generates the first line using the keywords given by users, and the other lines are generated according to the first line. \cite{yi2017generating,wang2016can,liu2018multi} use an Encoder-Decoder model with attention mechanism to generate poetry. Moreover, this model applies to the generation of Chinese Song iambics\cite{wang2016achinese}, which is a different genre in Chinese classical literature. To solve the problem of topic drift and improve the coherence of each line poetry, \cite{yan2016poet} proposes an iterative polishing schema and \cite{wang2016bchinese} propose a method which generates poems in a two-stage procedure, first they explicitly plan the sub-topic of each line, and then use both sub-topics and the preceding lines to generate the poem line by line. Most recently, researchers have put forward many new methods of Chinese poetry generation, such as employ memory network\cite{zhang2017flexible,yi2018chinese}, Variational AutoEncoder (VAE) \cite{yang2017generating} and reinforcement learning\cite{yi2018automatic}. Moreover, \cite{liu2019deep,guo2019jiuge} demonstrate the Chinese classical poetry generation system.

Further, generating poems from images is a more challenging task. \cite{liu2018beyond} proposes to generate English poems from
images in an end-to-end fashion. \cite{liu2018multi,cheng2018image,xu2018images} apply the extended Encoder-Decoder model with attention mechanism to generate classical Chinese poetry from images. However, there are still some defects in these approaches, such as the lack of image-poem paired dataset or the low quality of the paired dataset. Moreover, the information extracted implicitly from images is incomplete and uncontrollable, and hence the generated poems which based on the information will lack consistency with the given images. To resolve these problems, we propose a novel Chinese poetry generation approach to generate poems from images. To the best of our knowledge, this is the first method of generating poetry from images with non-parallel data. We also integrate the concrete and abstract information of images respectively into the model explicitly to address the issue of some important image information cannot be extracted accurately.

\section{Approaches}
Generating poems from images required that the information in images can be fully extracted and well reflected in poems. We propose a novel Chinese poetry generation approach to meet these two requirements, the framework is illustrated in Fig. \ref{fig2}. Our method is divided into two steps: image information extraction and poetry generation. We first extract the concrete keywords and abstract information from the given images. Then, our poetry generation model adds these concrete keywords to the poem in a fill-in-the-blank fashion, and the abstract information is explicitly integrated into the model using abstract information embedding.

Formally, The main aim of our approach is to generate a poem $P$ which consists of $L$ lines $\{l_1,l_2,...,l_L\}$ from a given image $I$. Our generation model takes two inputs: $T$ concrete keywords $K$ = \{$k_1,k_2,...,k_T$\} and $S$ abstract information labels $A$ = \{$a_1,a_2,...,a_S$\}, which are extracted from $I$. Then, the poem is generated line by line, when generating the $i$-th line $l_i$, the previous generated lines $l_{1:i-1}$ which represents the concatenation from $l_1$ to $l_{i-1}$, the $i$-th concrete keywords $k_i$ and the abstract information labels $A$ are used as input to our infilling-based model. Remarkably, the use of the previous lines $l_{1:i-1}$ is implicit, the use of concrete keywords $K$ and the abstract information labels $A$ are explicit. In this section, we will introduce the details of each component in our framework with the clue of concrete information and abstract information. The following subsections A, B will introduce the extraction and integration of concrete information and abstract information, respectively.

\subsection{Concrete Information}
The concrete information is the concrete physical objects in images, such as mountains, rivers. In our approach, the extracted keywords from images are treated as the concrete information of given images.

\paragraph{Information Extraction}
Given an image, we extract the concrete keywords by \emph{Clarifai}. The \emph{Clarifai} API offers image recognition as a service, it can recognize the object in the given image. Then, these modern Chinese keywords which recognized by \emph{Clarifai} should be transformed into ancient Chinese keywords, so we retrieve and expand all concrete keywords in \emph{ShiXueHanYing}, which a poetic phrase taxonomy built by Wenwei Liu in 1735. It contains 1016 manually picked themes and each theme consists of dozens of related phrases. In this way, we can extract the concrete information $K$ in the images and transform it into ancient Chinese related phrases that can be used in poetry generation directly. 

\begin{figure*}
\centering
\includegraphics[width=0.8\textwidth]{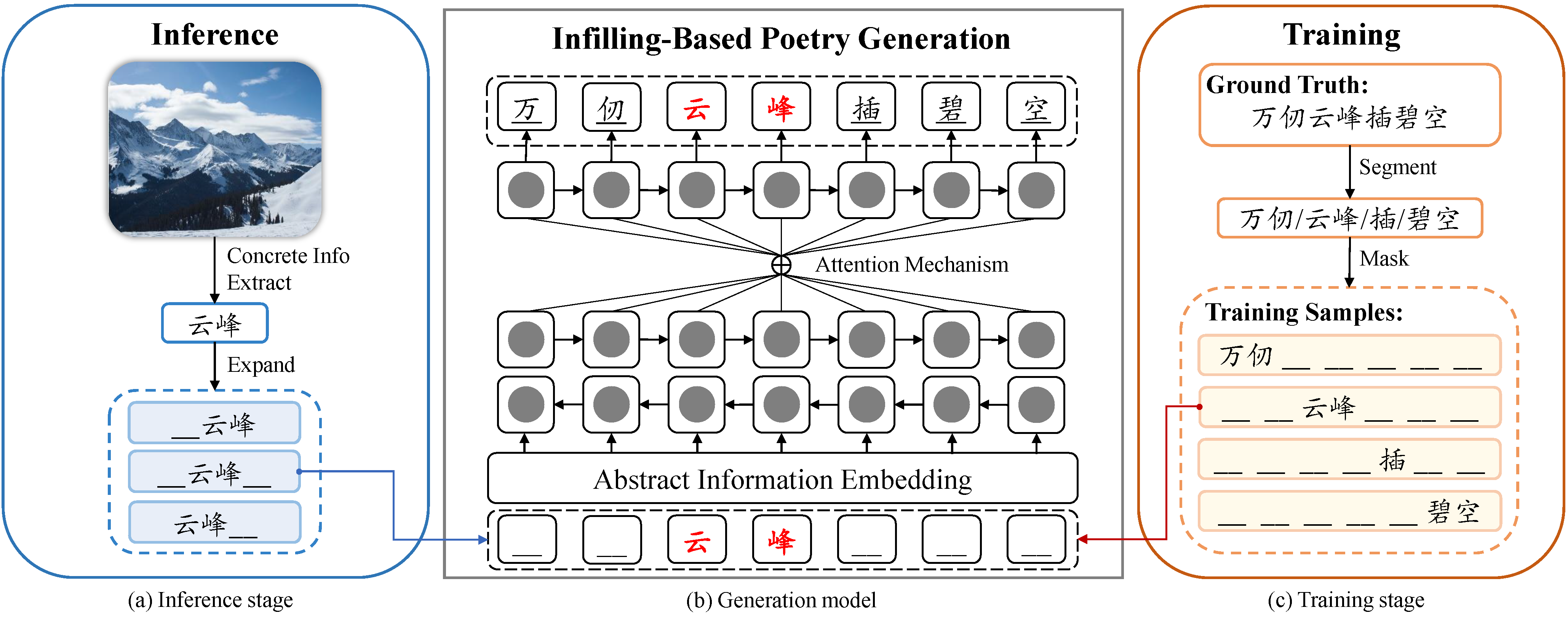}
\caption{An illustration of the infilling-based Chinese poetry generation model.}
\label{fig3}
\end{figure*}

\paragraph{Information Integration}
After extracting the concrete information from images, we propose an infilling-based poetry generation method to integrate the extracted concrete information into each line in the way of filling in the blank. As illustrated in part b of Fig. \ref{fig3}, The model is based on the Sequence-to-Sequence framework and generates poetry line by line, we use GRU\cite{cho2014learning} for bidirectional encoder and decoder. Different from the traditional seq2seq model, we use a fill-in-the-blank way to train our model, which can infill keywords into the lines of our poem. In previous methods, the Bi-GRU encodes the preceding generated lines $\{l_1,l_2,...,l_{i-1}\}$ when generating the $i$-th line of the poem. In our method, the input sequence is the preceding generated lines $\{l_1,l_2,...,l_{i-1}\}$ and the current $i$-th line which has masked the other words expect the keywords, the target sequence is the ground truth of current $i$-th line without masking. Formally, 
\begin{equation}
X = \{x_1, x_2, ... , x_N, \underline{y_1}, ..., \underline{y_{m-1}}, y_m, y_{m+1}, \underline{y_{m+2}}, ..., \underline{y_M}\}
\end{equation}
\begin{equation}
Y = \{y_1, y_2, ... , y_M\}
\end{equation}
where $X$ denotes the input sequence of our method and $Y$ denotes the target sequence, $\underline{y_m}$ donates the $m$-th word has been masked (replaced by underscore character $\_$ ), $N$ and $M$ represent the length of the input sequence and target sequence respectively.

For the encoder, let $e(x_n)$ represent the word embedding of character $x_n$, the $h_t$ represent the encoder hidden states which consist of the forward hidden vector $\{h_t\}$ and the backward hidden vector $\{h_t\}$ at the $t$-th step when encoding. That is,
\begin{equation}
\overrightarrow{h_t} = GRU(\overrightarrow{h_{t-1}} , e(x_t))
\end{equation}
\begin{equation}
\overleftarrow{h_t} = GRU(\overleftarrow{h_{t+1}} , e(x_t))
\end{equation}
\begin{equation}
h_t = [\overrightarrow{h_t};\overleftarrow{h_t}]
\end{equation}
for $t = 1, 2 ... T$ and $[\overrightarrow{h_t} ; \overleftarrow{h_t}]$ means the $t$-th hidden state $h_t$ is the concatenation of the forward hidden vector and the backward hidden vector of the Bi-GRU.

For the decoder, we use a GRU decoder with attention mechanism\cite{bahdanau2014neural}. For each step $m$ during decoding, the most probable output character $y_m$ is generated based on the previous generated character $y_{m-1}$, the internal status $s_m$ and the context vector $c_m$. Formally, we have:
\begin{equation}
y_m = \arg\max_y P(y|y_{m-1},s_m,c_m)
\end{equation}
where $s_m$ is updated by
\begin{equation}
s_t = f(y_{m-1},s_{m-1},c_{m-1})
\end{equation}
$f(.)$ is the activation function of GRU and $c_m$ is a weighted sum of all encoder hidden states $h_1, h_2, ... ,h_T$ which calculated by attention mechanism\cite{bahdanau2014neural}.
\begin{equation}
c_m = \Attention(s_{m-1},h_{1:T})
\end{equation}
It is noteworthy that the framework we proposed is using non-parallel data for training. For the concrete keywords, we use the original words of the poem from the poetry dataset to train our infilling-based poetry generation model. Next, we will discuss the training and inference stages respectively.

During training, we first construct a poetry dataset in the form of filling in blanks to fit our task. As shown in part c of Fig. \ref{fig3}, the ground truth in our dataset is first segmented word by word, then we keep one word and mask the others, and the training samples are the whole sentence after being masked. In this way, the reserved words are from ground truth itself rather than relating to images so that the image-poem pair data is not required during training.

During inference, as shown in the part a of Fig. \ref{fig3}, we first use \emph{Clarifai} to extract the concrete keywords and use the \emph{ShiXueHanYing} to retrieve the classical literary form of all keywords. Then we give these words to our generation model, the concrete keywords are integrated into the model using a fill-in-the-blank way which used during training. For example, the keyword $k_i$ is considered in three ways: $k_i \_$, $\_ k_i \_$, and $\_ k_i$ which represent the position of $k_i$ is at the front, middle and back of a line. The model will consider these three forms to fill the keywords into a line, then choose the best sentence from the candidate sentences.

\begin{figure*}
\centering
\includegraphics[width=0.82\textwidth]{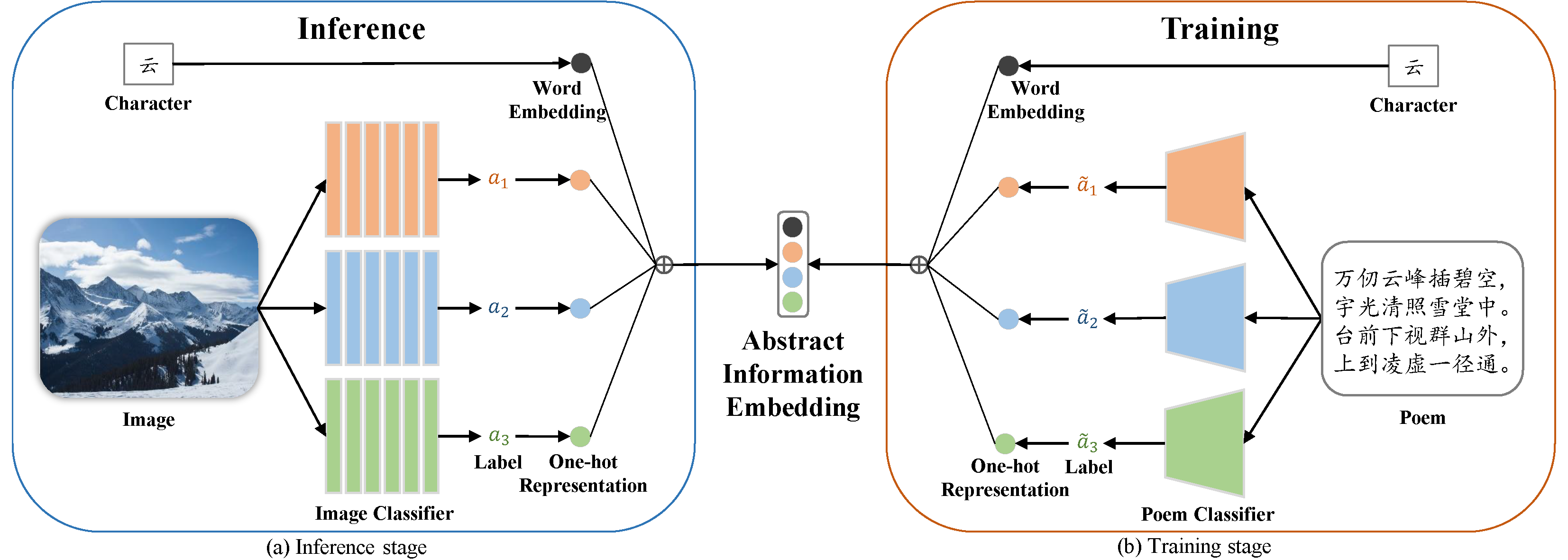}
\caption{An illustration of the abstract information embedding.}
\label{fig4}
\end{figure*}

\subsection{Abstract Information}
The abstract information is the implied background information and people's common sense, such as season information. It is noteworthy that the abstract information is hard to be recognized by machines but easy to be recognized by people. For this reason, it is important for the model to recognize the right abstract information and reflected it in generated poems, otherwise, the quality of the generated poems will be very poor. for example, if the generated poem in Fig. \ref{fig2} contains some phrases of spring, but the abstract information of the given image is winter, this will lead to poor results. Therefore, in this work, we consider three kinds of abstract information: season information, time information and weather information, each contains several categories. Season information is divided into five categories: spring, summer, autumn, winter and other, they can represent the seasons of the images. Similarly, the categories of time information are daytime, nightfall, night and other; the categories of weather information are sunshine, cloudy, rainy, snowy and other. Especially, the ``other" category means that we cannot judge it from the image.
\paragraph{Information Extraction}
As we mentioned above, we divide abstract information $A$ into three categories. As shown in the left side of Fig. \ref{fig4}, first we use three image classification model to predict the abstract category from these three aspects. The classification model is based on the GoogleNet which is pre-trained on ImageNet and fine-tuned on the target dataset. In order to do this, we build three target datasets corresponding to the three categories of abstract information manually. Especially, if the probability of the top-1 result is less than 80\% during inference, we set the category to ``other", which means that we cannot judge it from the image. With these three image classification models, we get three labels $a_1$, $a_2$ and $a_3$ corresponding to three aspects of the abstract information. Formally, the abstract information $A$ is predicted from the given images using three CNN models, which is
\begin{equation}
A = {a_1, a_2, a_3}
\end{equation}
The accuracy of these three CNN predictors are 97.75\%, 96.92\% and 90.63\%, which is trained by the season, time, weather dataset respectively.

\paragraph{Information Integration}
After extracting the abstract information from images, We propose an abstract information embedding to concatenate them into generated poems when encoding. Let $a_1$, $a_2$ and $a_3$ donate the three labels corresponding to three aspects of the abstract information, then we transform them into one-hot representation $v_1$, $v_2$ and $v_3$, respectively. To keep the generated poem and one-hot representation consistent, we put forward an abstract information embedding. We concatenate these three one-hot representations and the word embedding vector $e(x_n)$, then feed the concatenated vector into the encoder model substitute for the general word embedding. That is,
\begin{equation}
v_i = \onehot(a_i)
\end{equation}
\begin{equation}
\tilde{e}(x_n) = [v_1; v_2; v_3; e(x_n)]
\end{equation}
where the range of $i$ is from one to three, $\onehot(.)$ transform the labels into one-hot representation and the $\tilde{e}(x_n)$ represents the abstract information embedding of character $x_n$.

Specially, we also use non-parallel data in this component instead of image-poem paired data, so that the data using in the training stage is different from the inference stage. Next, we will introduce these two stages separately.

During training, as shown in part b of Fig. \ref{fig4}, we first classify our poems dataset according to three aspects of abstract information. After this, each poem in the dataset has three extra labels, then we convert these labels into three one-hot vectors. These vectors are applied in the abstract information embedding and replace the real abstract information extracted from the images.

During inference, as shown in the part a of Fig. \ref{fig4}, we use the image classification model to predict the abstract information, then transform the predicted labels into one-hot representation and concatenate it with word embedding vector. Then we put this vector in the model instead of the general word embedding. The extracted labels represent the abstract information of the given image, and these labels $\{a_1, a_2, a_3\}$ replace the label of poetry which using in training. Thus, our model can generate poetry using the information from the images and is not required the image-poem pair data for training. 

\section{Experiments}

\subsection{Dataset}
In this paper, we use non-parallel data rather than the image-poem paired dataset. For reaching the better effect, we build a large poetry corpus, in which the data is the form of filling-in-the-blank. The infilling-based corpus contains 3300709 Chinese poems which are masked some words randomly. We randomly select 80\% of the corpus as the training set, 10\% as the validation set and 10\% as the testing set. The vocabulary size of this poetry corpus is 5297 and the corpus is used to train the infilling-based poetry generation model. 

Furthermore, we classify all poems of the poetry corpus above to build a poetry classification dataset. First, we choose several representative words from each category, the appearance of these words often determines which category a poem belongs to, For example, the appearance of the word ``Lotus flower" often means that the poem is describing summer. Then we retrieve the poems that contain these words in the poetry corpus and build a small-scale poetry classification dataset manually using the retrieval results. Using this small-scale dataset, we train a poetry classification model based on BERT\cite{devlin2018bert} and construct a poetry classification dataset, it has the same size as the previous poetry corpus and has three labels attached which correspond to the three categories of abstract information. These three labels are used as abstract information when training the abstract information embedding. 

Moreover, we construct three image classification datasets manually. Each dataset corresponds to an aspect of abstract information and each dataset contains 9600 images, 1200 of which are used for verification, 400 of which are used for test and the rest for training. These image datasets are used to train the image classification models which are used to predict the abstract category from images during inference. 

\subsection{Training Details}
For the poetry generation model, the dimension of the encoder and decoder hidden states is 1024, the beam search size is 5, the dimensions of character embeddings are 1024 and the dimensions of each one-hot abstract information representation is 5. Hence the dimension of abstract information embeddings is 5 $\times$ 3 + 1024 = 1039. Our model is trained with the mini-batches setting to 256 and the Adam optimizer is used for stochastic gradient descent. We set the learning rate to 0.001 and the dropout strategy with dropout rate 0.5 is used to avoid overfitting during training. Moreover, we use gradient clip and weight decay to improve our model. The model is selected according to the cross-entropy loss on validation set.

For the poetry classification model, we use a base version BERT which has 12 layers, 12 heads, and 110M parameters. The model is pre-trained on the Chinese dataset and fine-turn on our poetry classification dataset. For the image classification model, we use a pre-trained Inception-v3 model and fine-tune on our image classification dataset. The batch size is set to 32 and the learning rate is set to 0.001. For each image, 3 abstract information labels are predicted and 10-20 keywords are extracted.

\begin{table*}
\centering
\caption{Human evaluation results of all methods}
\begin{tabu}{|c|c|c|c|c|c|c|}
\hline 
\textbf{Methods}&\textbf{Poetiness}&\textbf{Fluency}&\textbf{Coherence}&\textbf{Meaning}&\textbf{Consistency}&\textbf{Average}\\
\hline 
RNNPG-A&6.75&6.61&6.58&6.32&5.89&6.43\\
\hline 
RNNPG-H&6.78&6.66&6.47&6.41&5.97&6.46\\
\hline 
PPG-R&6.85&6.40&6.59&6.01&5.46&6.26\\
\hline 
PPG-H&6.99&6.55&6.85&6.55&6.11&6.61\\
\hline 
WMPG-R&7.98&7.67&7.67&7.16&6.46&7.39\\
\hline 
WMPG-H&8.06&7.68&7.76&7.45&6.85&7.56\\
\hline 
BFPG&7.61&7.39&7.51&7.24&6.86&7.32\\
\tabucline[0.7pt]{1-7}
IPG(full)&\textbf{8.31}&\textbf{7.91}&\textbf{8.03}&\textbf{7.83}&\textbf{7.76}&\textbf{7.97}\\
\hline 
IPG(w/o concrete info)&7.62&7.65&7.70&7.45&7.21&7.53\\
\hline 
IPG(w/o abstract info)&7.86&7.73&7.62&7.48&6.82&7.50\\
\hline 
\end{tabu}
\end{table*}

\subsection{Evaluation Design}
\paragraph{Automatic Evaluation Metrics}
Up to now, the accurate automatic evaluation of text generation is still very difficult. The previous evaluation metrics of poetry generation include BLEU, ROUGE, and perplexity\cite{he2012generating,zhang2014chinese}. However, these metrics are not very effective for poetry generation, and it has been proved that these metrics correlate very weakly with human judgements\cite{liu2016not}. Moreover, the consistency between poems and images is an important evaluation metric when generating poems from images. In this work, we propose a new automatic metric INFO to evaluate whether the basic information from images is well reflected in the poems, we classify the generated poems from three aspects automatically using the well-trained neural network models. then we check whether these categories match the categories of the given image, and we take the matching degree as a measure, the INFO score is the matching correctness rate between poems and images for each method. The test images are classified manually.
\paragraph{Human Evaluation Metrics}
Different from ordinary sentences, poetry is a literary form with a concise language and abundant creativity. Therefore, human evaluation is the major metric for this task. As the previous works\cite{zhang2014chinese,wang2016achinese,xu2018images} did, our human evaluators will evaluate the generated poems from these four standards:

\begin{itemize}
\item \textbf{Poeticness}: Does the poem follow the rhyme and tone regulations?
\item \textbf{Fluency}: Does the poem read smoothly and fluently?
\item \textbf{Coherence}: Is the poem coherent across lines?
\item \textbf{Meaning}: Does the poem have a reasonable meaning and artistic conception?
\item \textbf{Consistency}: Dose the poem match the given image well?
\end{itemize}

\subsection{Baselines}
We implemented some works on generating Chinese classical poetry from images as baselines. Because the input of some methods is keywords, we directly give the keywords which extract from the given images to these baselines. The baselines including:

\textbf{RNNPG}: A Chinese poetry generation method based on the recurrent neural networks\cite{zhang2014chinese}. The first line corresponds to the keywords and generates using templates, the generation of the rest lines are based on the previous lines. Furthermore, we implement RNNPG-A and RNNPG-H, the former uses all keywords to generate the first line and the latter only uses a part of keywords selected manually. 

\textbf{PPG}:A Chinese poetry generation method with planing based on an attention encoder-decoder framework\cite{wang2016bchinese}. The model explicitly plans a sub-topic keyword of each line. Specifically, we have two ways to implement the model: PPG-R and PPG-H. For PPG-R, the model randomly selects four keywords from the candidate keywords. For PPG-H, these four keywords for four lines are chosen by the human.

\textbf{WMPG}: A Chinese poetry generation method with working memory framework\cite{yi2018chinese}. The model maintains the keywords in the memory explicitly, and the generation model learns the most relevant information in memory when generating each line. Similarly, the keywords are given to the model in two ways: select randomly and manually. Therefore, we have two ways to implement the model: WMPG-R and WMPG-H.

\textbf{BFPG}: A Chinese poetry generation from images method based on a backward language model and a forward language model\cite{liu2018multi}. The keywords extracted from images automatically are first reversed and fed into the backward language model to generate the first half part of the line, then the result is used to generate the whole line by the forward language model. The method can generate poems from images automatically, hence we give the image to the model directly.


\subsection{Model Variants}
In addition to the baselines, we propose two variants of our model to evaluate whether each component of our model is effective and necessary, then compare these variants with the baselines and the full model.

\textbf{IPG(full)}: The propose infilling-based Chinese poetry generation model which replaces the general word embedding with our abstract information embedding.

\textbf{IPG(w/o concrete info)}: Based on IPG, the concrete information is removed by using the general sequence-to-sequence Chinese poetry generation model instead of filling the keywords into the poems. The abstract information embedding is employed as normal.

\textbf{IPG(w/o abstract info)}: Based on IPG, the abstract information is removed by using the general word embedding instead of the abstract information embedding, and the other part of the infilling-based Chinese poetry generation model is same as the full model.

\begin{table}
\centering
\caption{Automatic evaluation results of all methods}
\begin{tabu}{|c|c|c|c|}
\hline 
\textbf{Methods}&\textbf{INFO}&\textbf{Methods}&\textbf{INFO}\\
\hline 
RNNPG-A&23.90\%&WMPG-H&40.90\%\\
\hline 
RNNPG-H&37.40\%&BFPG&38.10\%\\
\hline 
PPG-R&24.80\%&IPG(full)&\textbf{88.69}\%\\
\hline 
PPG-H&30.90\%&IPG(w/o concrete info)&77.80\%\\
\hline 
WMPG-R&38.70\%&IPG(w/o abstract info)&39.09\%\\
\hline 
\end{tabu}
\end{table}


\subsection{Evaluation Results}
\paragraph{Automatic Evaluation Results}
To evaluate the consistency of the abstract information between the generated poems and given images automatically, we conduct the INFO evaluation metric by comparing the abstract labels of the poems and images.

The INFO results are shown in Table 2, the INFO score is the matching correctness rate between poems and images for each method. The higher score of INFO represents that the generated poem is more consistent with the given image in terms of abstract information. From the table, we can see that the methods (RNNPG, PPG, WMPG, BFPG) which based on the keyword have a low score in INFO metrics, and their INFO score is far worse than our method (IPG), this indicates the concrete keywords cannot express the abstract information of given images well. Moreover, the two variants (IPG w/o concrete info and w/o abstract info) of our model differ greatly in INFO score, the score of IPG (w/o abstract info) is similar to WMPG and BFPG, but notable lower than IPG (w/o concrete info), it is sufficient to show that the abstract information embedding component is very effective for integrating the abstract information from images. One can observer IPG (full) outperforms the two variants, which shows that the concrete keyword also has an influence in the INFO score because some keywords imply abstract information. 
\paragraph{Human Evaluation Results}
Moreover, we conduct human evaluation to compare the baselines and variants. For each method, we collect 10 generated poems from 10 images respectively. We invite 20 evaluators who are proficient in Chinese literature to rate each poem with a score from 1 to 10 according to our evaluation standards.

Table 1 displays the results of human evaluation. Firstly, the results demonstrate our approach IPG (full) outperforms all baselines and variants, the ``Consistency" scores show that IPG can improve the consistency of the generated poems and given images, the other metrics indicate that our approach can generate better poems from images without losing rhythm, fluency, coherence, and meaning. Then we can obverse that the baselines which the input keywords are selected manually outperform the contrast methods which inputs the random keywords or all keywords, including RNNPG-A and RNNPG-H, PPG-R and PPG-H, WMPG-R and WMPG-H. This demonstrates that the keywords can affect the quality of the poems and the consistency between the poems and images. Also, the score of IPG (w/o abstract info) is better than PPG-H and close to WMPG-H, which indicates that our method can integrate the keywords explicitly to improve consistency while ensuring quality. 

Comparing these two variants, it is worth noting that the evaluation scores are close except ``Consistency" and the ``Consistency" score of IPG (w/o concrete info) is higher than IPG (w/o abstract info) and all baselines. This proves that abstract information has more influence on consistency than keywords. Moreover, these two variants degrade the performance of our approach IPG (full), which indicates that the concrete information and abstract information cooperate to generate high-quality poems consistent with given images.

\begin{figure*}
\centering
\includegraphics[width=0.85\textwidth]{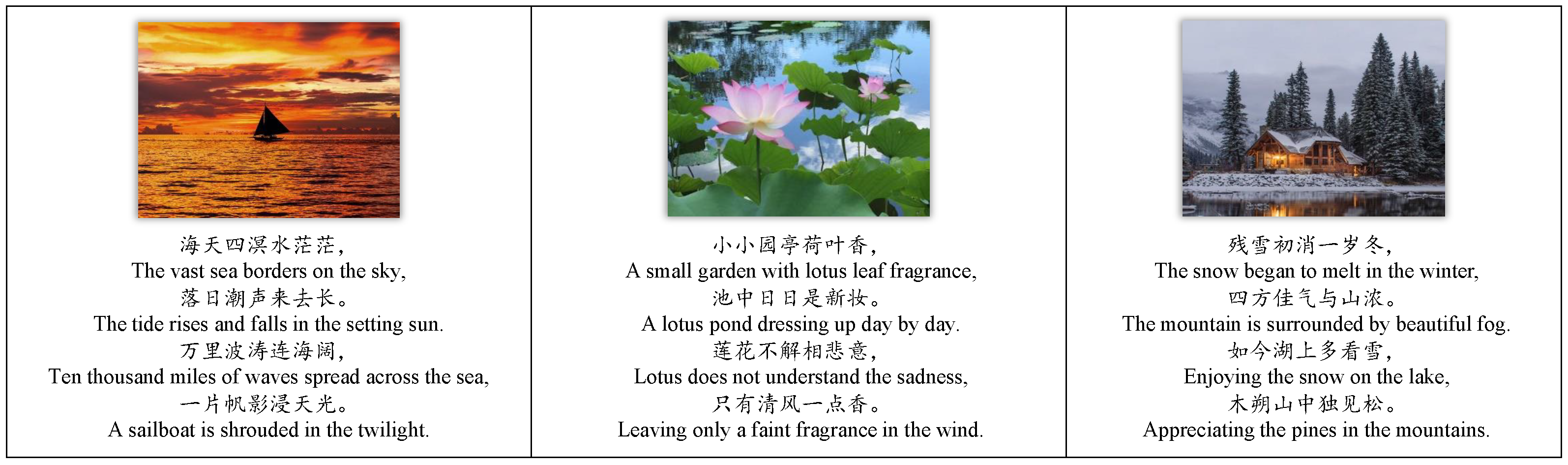}
\caption{Example of poems generated from images by our approach IPG.}
\label{fig5}
\end{figure*}

\subsection{Examples}
Fig. \ref{fig5} shows example poems generated by our IPG framework, and the given images are above the generated poems. From these examples, we can see that the poems generated by the IPG framework can capture the concrete information as well as the abstract information from the given images, and keep the strict metrical patterns of poetry. Especially, the concrete information is well reflected in poems, such as ``nightfall" in the left example, ``summer" in the middle example and ``winter" in the right example. 

\section{Conclusion}
In this paper, we propose a new approach to generate classical Chinese poetry from images. Given an image, we first extract the information of the image from two aspects: the Concrete information and the Abstract information. Then, the infilling-based poetry generation model integrates these concrete keywords into each line and the abstract information embedding integrates the labels of the abstract information. Furthermore, we use a novel training strategy that using non-parallel data during training so that the image-poem paired datasets are not required. Comparison experiment and ablation study demonstrate that our framework outperforms all baselines and variants. The automatic and human evaluation shows that our approach can capture the concrete information and the abstract information from images without losing the quality of generated poems.

\bibliographystyle{IEEEtran}
\bibliography{reference}
\end{document}